\documentclass[lettersize,journal]{IEEEtran}
\usepackage{amsmath,amsfonts}
\usepackage{array}
\usepackage{textcomp}
\usepackage{stfloats}
\usepackage{url}
\usepackage{verbatim}
\usepackage{graphicx}
\def\BibTeX{{\rm B\kern-.05em{\sc i\kern-.025em b}\kern-.08em
		T\kern-.1667em\lower.7ex\hbox{E}\kern-.125emX}}
\usepackage{balance}
\usepackage{xcolor}
\usepackage{algorithm}
\usepackage{algpseudocode}
\usepackage{booktabs}
\usepackage{caption}
\usepackage{subcaption}
\usepackage{ amssymb }
\usepackage{multirow}

\begin{document}
	
	
	\title{CIDGIKc: Distance-Geometric Inverse Kinematics for Continuum Robots}
	\author{Hanna Jiamei Zhang$^a$, Matthew Giamou$^b$, Filip Mari\'c$^b$, Jonathan Kelly$^b$, and Jessica Burgner-Kahrs$^a$
		\thanks{Manuscript received [?] 2023; Revised [?] 2023; Accepted [?] 2023; This paper was recommended for publication by Editor [?] upon evaluation of the Associate Editor and Reviewers' comments.\\We acknowledge the support of the Natural Sciences and Engineering Research Council of Canada (NSERC) [RGPIN-2019-04846]. \\ $^a$Continuum Robotics Laboratory, Department of Mathematical \& Computational Sciences, University of Toronto, Canada. \\ $^b$ Space \& Terrestrial Autonomous Robotic System Laboratory, Institute for Aerospace Studies, University of Toronto, Canada}}
	
	%
	
	
	\maketitle
	
	\begin{abstract}
		The small size, high dexterity, and intrinsic compliance of continuum robots (CRs) make them well suited for constrained environments. 
		Solving the inverse kinematics (IK), that is finding robot joint configurations that satisfy desired position or pose queries, is a fundamental challenge in motion planning, control, and calibration for any robot structure. For CRs, the need to avoid obstacles in tightly confined workspaces greatly complicates the search for feasible IK solutions. Without an accurate initialization or multiple re-starts, existing algorithms often fail to find a solution. 
		We present \verb+CIDGIKc+ (Convex Iteration for Distance-Geometric Inverse Kinematics for Continuum Robots), an algorithm that solves these nonconvex feasibility problems with a sequence of semidefinite programs whose objectives are designed to encourage low-rank minimizers. 
		\verb+CIDGIKc+ is enabled by a novel distance-geometric parameterization of constant curvature segment geometry for CRs with extensible segments. 
		The resulting IK formulation involves only quadratic expressions and can efficiently incorporate a large number of collision avoidance constraints. 
		Our experimental results demonstrate \textgreater98\% solve success rates within complex, highly cluttered environments which existing algorithms cannot account for. 
	\end{abstract}
	
	
	\section{Introduction}
	\label{sec:intro}
	\IEEEPARstart{C}{ontinuum} robots (CRs) have attracted attention for their dexterity, intrinsic compliance, and miniaturizability. 
	These qualities enable their deployment in confined environments with sensitive structures such as those encountered in minimally invasive surgery and industrial inspection tasks \cite{burgner2015continuum}. 
	Solving  the inverse kinematics (IK) problem of CRs with a high success rate in obstacle-laden workspaces remains an open challenge. 
	To our knowledge, there is no IK solver for multi-segment, extensible CRs able to perform at scale in heavily cluttered environments. 
	
	There is an emerging class of IK techniques for serial manipulators that leverages alternative parameterizations based on the distances between control points affixed to a robot~\cite{blanchini_convex_2017, le2019kinematics, maric2020serial}. 
	In return for a higher dimensional problem, this distance-geometric (DG) perspective elegantly describes the robot workspace with simple pairwise distance constraints. 
	Inspired by the DG approaches of \cite{giamou2022cidgik} and \cite{maric2021riemannian} for serial manipulators, we leverage semidefinite programming (SDP) relaxations for DG problems to develop \texttt{CIDGIKc}, a novel IK solver for extensible multi-segment continuum robots. 
	With the flexibility offered by this formulation, it is possible to solve IK problems considering position, orientation, and full pose constraints of the end effector. Additionally, the DG representation can naturally incorporate numerous spherical obstacles and planar workspace constraints. 
	Finally, we provide an open source Python implementation of \texttt{CIDGIKc}.
	\footnote{\url{https://github.com/ContinuumRoboticsLab/CIDGIKc}}
	\begin{figure}[!t]
		\centering
		\includegraphics[width=0.49\textwidth]{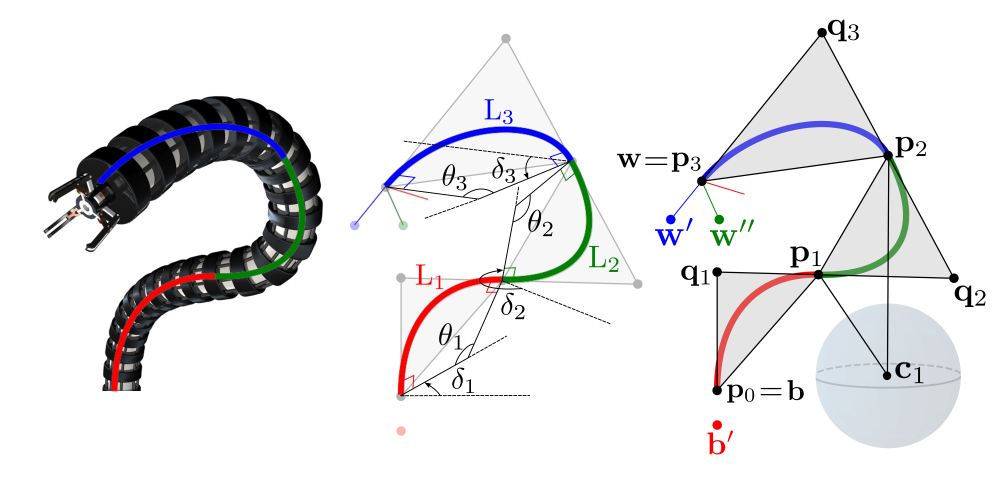}\\
		\vspace{1mm}
		\includegraphics[width=0.49\textwidth]{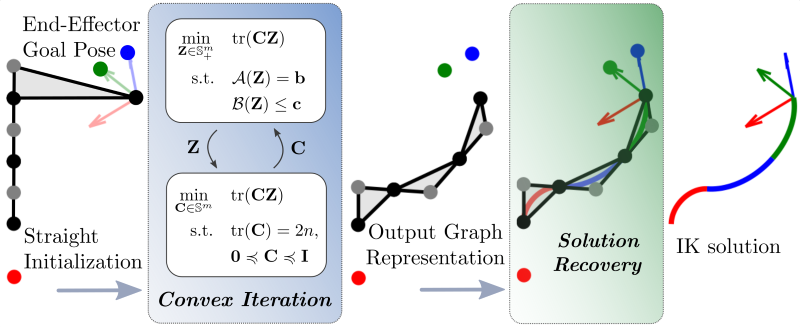}
		\caption{We introduce a graph-based IK solver for continuum robots based on low-rank convex optimization. Top: 3-segment spatial CR (left) overlaid with its body shape representation, (middle) given a constant curvature segment shape representation, and (right) converted to our graphical representation with a spherical obstacle avoidance constraint. Bottom: \texttt{CIDGIKc} pipeline for solving inverse kinematics of continuum robots.}
		\label{fig:3d_model_and_pipeline}
	\end{figure}
	
	\section{Related Work}
	\label{sec:rel_works}
	In this section, we briefly review the state of art in IK for CRs, distance geometry, and semidefinite optimization. 
	\subsection{Continuum Robot Inverse Kinematics}
	Modelling the forward kinematics (FK) of CRs requires a simplification of their continuously bending, infinite degree of freedom (DoF) structure. 
	One widely-used simplification treats each independently bending segment as a constant curvature (CC) circular arc~\cite{webster2010design}. 
	To consider CRs with variable curvature (VC) segments, more advanced methods involving elliptic integrals, Cosserat rod theory, or lumped models are required~\cite{rao2021model}. 
	With VC models, more realistic shape results come at the cost of increased computational effort, limiting their use in practical IK problems with time and resource constraints.
	
	A popular set of approaches is to incrementally drive a CR's end-effector (EE) to a desired pose using the exact or approximate inverse Jacobian. These methods involve costly inverse Jacobian calculations, are susceptible to convergence issues, prone to singularity/numerical stability issues, require an accurate initialization, and typically have no formal guarantees. Related are optimization methods which leverage accurate forward kinematics models to solve IK~\cite{singh2017performances}. However, these objective functions are non-convex, and constraints on joint angles involve trigonometric functions, both of which are very difficult for local and global solvers alike to handle. For the remainder of this paper, we limit our discussion to IK solvers which leverage the CC assumption. If each segment is represented by a rigid link connecting its base and tip and per-section endpoints are known a priori, then analytical solutions can be employed~\cite{neppalli2009closed}.
	In the more common case where endpoints are unknown, they can be estimated through a procedural guess-and-check style of endpoint placement to resolve the redundancy. For problems with prescribed segment lengths and EE orientation, the method in \cite{neppalli2009closed} has no solution guarantees. 
	
	Heuristic iterative methods have shown promising results. 
	The FABRIKc algorithm \cite{zhang2018fabrikc}, inspired by its serial arm counterpart FABRIK \cite{aristidou2011fabrik}, leverages a kinematic representation where each CC segment is replaced by two rigid links and a joint. 
	FABRIKc is limited to EE goals with five DoF where roll is not specified. 
	FABRIKx \cite{kolpashchikov2022fabrikx} extends FABRIKc to handle variable curvature fixed-length segment CRs. 
	The CRRIK algorithm \cite{wu2022crrik} also builds upon FABRIKc to offer solves that are 2.02 times faster through a modification of the joint placement strategy, which is able to avoid large joint position offsets.
	CRRIK includes a check and repositioning of segments based on collisions on the backwards iteration to avoid obstacles. 
	For these methods, only constant segment lengths can be considered. 
	Due to its iterative obstacle collision avoidance scheme, CRRIK solve times are expected to scale poorly in environments where the number of obstacles is very large however such a study is not presented in \cite{wu2022crrik}.  
	The SLInKi solver \cite{chiang2021slinki} leverages state lattice search, commonly used in path-finding problems, to resolve redundancy resulting from multiple extensible bending segments. 
	A component of SLInKi is exFABRIKc, which modifies the forward reaching step of FABRIKc and adjusts the virtual link lengths to enable arc length variation.
	SLInKi combines state lattice search and exFABRIKc for high speed searching and accurate end pose respectively.
	The method in \cite{garriga2019kinematics} leverages quaternions to derive closed-form solutions for CRs with up to three extensible segments. 
	The approach of \cite{kim2019inverse} leverages DH parameters to solve IK for tendon-driven robots in particular. 
	With the exception of CRRIK, all solvers surveyed thus far lack an obstacle avoidance strategy. 
	
	In sum, the following open challenges remain: 1) handling extensible segments, 2) guaranteeing solutions i.e., algorithmic success, 3) incorporating arbitrarily many obstacle avoidance constraints, and 4) specifying a variety of EE goals (e.g., positions or full 6 DoF poses).
	
	\subsection{Distance Geometry}
	Many problems can be expressed with Euclidean distance geometry using points and their relative pairwise distances. 
	These distance geometry problems (DGPs) appear in a variety of applications including sensor network localization, molecular conformation computation, microphone calibration, and indoor acoustic localization~\cite{dokmanic2015euclidean}. 
	A distance-geometric formulation of IK for serial manipulators was first presented in \cite{porta2005inverse} for 6 DoF manipulators, and was solved for more general cases with a complete but computationally expensive branch-and-prune solver in \cite{porta2005branch}. 
	Recent approaches solve DGP formulations of IK for redundant serial manipulators with SDP relaxations \cite{giamou2022cidgik} and Riemannian optimization \cite{maric2021riemannian} for superior performance when handling numerous obstacles compared to state-of-the-art methods.
	
	
	\section{Problem Formulation}
	\label{sec:prob_form}
	
	We present a novel graph-based kinematic model of multi-segment extensible CC robots to express IK as a quadratically-constrained quadratic program (QCQP). 
	The model treats each independently-actuatable robot segment as an isosceles triangle described with distance constraints. 
	This enables the formulation of IK for CR with workspace constraints as convex feasibility problems whose low-rank feasible points provide exact IK solutions. 
	Inspired by the success of \cite{giamou2022cidgik} for serial manipulators, we apply the convex iteration method of \cite{dattorro2010convex} to our novel graph representation for continuum robots. Our method address open challenges 1), 3), and 4). We demonstrate progress made towards 2) through extensive experimentation.
	
	This section contains a detailed description of our distance geometry-inspired graph kinematic model of CC extensible segment continuum robots. Boldface lower and upper case letters (e.g., $\mathbf{x}$ and $\mathbf{P}$) represent vectors and matrices respectively. We write $\mathbf{I}_n$ for the $n\times n$ identity matrix, omitting the subscript when it is clear from context. The space of $n\times n$ symmetric and symmetric positive semidefinite (PSD) matrices are denoted $\mathbb{S}^n$ and $\mathbb{S}^n_+$, respectively. We denote the indices ${0,1\dots,n}$ as $[n]$ for any $n\in\mathbb{N}$. The Euclidean vector norm is denoted $||\cdot||$.
	
	\subsection{Inverse Kinematics}
	IK is the problem of finding, for a given EE position or pose goal, robot joint angles referred to as configuration space $\mathbf{\Theta} \in \mathbf{\mathcal{C}}$ that satisfy $F(\mathbf{\Theta}) = \mathbf{W} \in \mathbf{\mathcal{W}}$, where $F: \mathbf{\mathcal{C}} \rightarrow \mathbf{\mathcal{W}}$ is the trigonometric \textit{forward} kinematics function that maps joint angles in the configuration space $\mathbf{\mathcal{C}}$ to EE positions or poses in the workspace $\mathbf{\mathcal{W}}$. When a closed form solution $\mathbf{\Theta} = F^{-1}(\mathbf{W})$ is not feasible, which is often the case for redundant manipulators, numerical methods are traditionally used to solve optimization-based formulations of IK. 
	
	\textit{Problem 1} (Inverse Kinematics). Given a robot's forward kinematics $F: \mathbf{\mathcal{C}} \rightarrow  \mathbf{\mathcal{W}}$ and a desired EE position or pose $\mathbf{W} \in \mathbf{\mathcal{W}}$, find the configurations $\mathbf{\Theta}$  that solve
	\begin{align}
		\min_{\mathbf{\Theta} \in \mathbf{\mathcal{C}}} ||F(\mathbf{\Theta})-\mathbf{W}||^2
	\end{align}
	\subsection{Graph Representation}
	\label{subsec:dg_model}
	We consider a general multi-segment continuum robot with $n$ independently actuatable CC segments where $t$ denotes the segment index $t=[n]$ and higher $t$ correspond to more distal segments. To describe the $t^{\text{th}}$ CC segment, traditionally the following configuration space parameters have been used: 
	segment length $L_t \in [L_{t,min}, L_{t,max}]$, bending angle of the segment in the bending plane  $\theta_t$ ($\theta_t=0$ corresponds to a straight segment), and rotation of the bending plane $\delta_t$.
	Our proposed kinematic model uses the simplifying CC assumption. However, we abandon the notion of arc parameters in favour of control points embedded in ${\rm I\!R}^{d}$ where $d = 2$ and $d = 3$ for planar and spatial robots, respectively. 
	The positions of these points are subject to distance constraints that describe feasible CC segments, and linear constraints that enforce the smoothness of the overall shape of a multi-segment CR.
	The points are strategically fixed along the robot body in relation to each individual CC segment $t$: $\mathbf{p}_t$, $\mathbf{p}_{t-1}$ are positions of the tip and base respectively, and $\mathbf{q}_t$ is the position of the ``virtual joint" which is located at the intersection of the base and tip CC arc tangents. 
	These three points form an isosceles triangle which represent the $t^{\text{th}}$ independently bending CC segment. 
	
\begin{figure}[!b]
		\centering
		\includegraphics[width=0.44\textwidth]{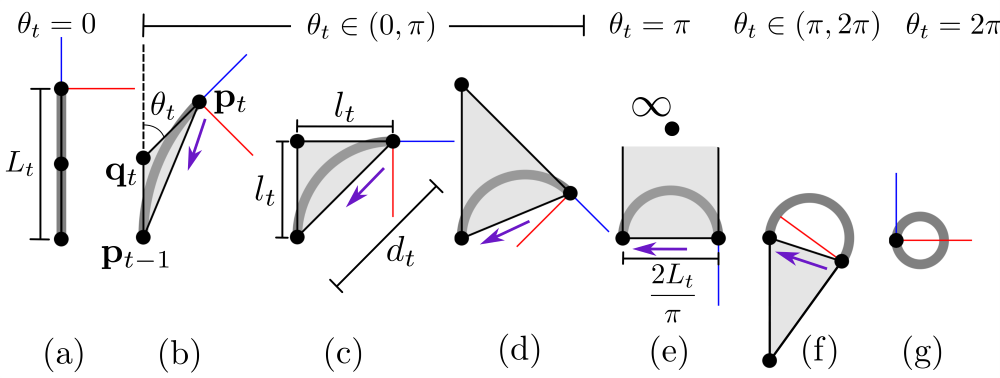}
		\caption{General and special bending cases for a single planar fixed-length segment $t$ modelled with the constant curvature assumption. Each case is overlaid with its distance-geometric description.}
		\label{fig:2d_model}
	\end{figure}

	The different bend configurations of a fixed-length segment using this model are shown in Fig. \ref{fig:2d_model}. 
	We consider segments bending according to the cases shown in Fig. \ref{fig:2d_model}a) through Fig. \ref{fig:2d_model}d), that is $\theta_t \in [0, \pi)$. 
	Obtaining the correct positions of all robot body points $\mathbf{p}_t$ and $\mathbf{q}_t$ for each segment $t$ is equivalent to obtaining the underlying CC arcs configuration $\mathbf{\Theta} \in \mathbf{\mathcal{C}}\subseteq \mathbf{\mathcal{T}}^N$, where $\mathbf{\mathcal{T}}^N$ is the $N$-dimensional torus. 
	
	Our graph representation, shown in Fig. \ref{fig:3d_model_and_pipeline}, uses a set of fixed points to specify the goal position or pose of a robot's EE and base with $\mathbf{W} = \begin{bmatrix}\mathbf{b}&\mathbf{b}^\prime&\mathbf{w}&\mathbf{w}^{\prime} & \mathbf{w}^{\prime\prime}\end{bmatrix} \in {\rm I\!R}^{d\times k}$ collectively referred to as \textit{anchors}. There are $k$ total anchor points noted $\mathbf{a}_k$ with $3\leq k\leq5$. Three anchors must be the position and orientation specifying points of the robot base and EE respectively $\mathbf{b} \stackrel{\rm def}{=} \mathbf{p}_0$, $\mathbf{b}^\prime$, and $\mathbf{w} \stackrel{\rm def}{=} \mathbf{p}_n$. All other points composing the robot body, $\mathbf{p}_t$ and $\mathbf{q}_t$, are considered \textit{unanchored} points. There are $j$ total unanchored/robot body points denoted $\mathbf{u}_j$ with $j=2(n-1)+1$. In order to form a graph describing a CC CR representation let $\mathcal{V}_{\mathbf{j}} = [j]$ and $\mathcal{V}_{\mathbf{k}}=[k]$ be index sets for vertices representing the unanchored and anchored points respectively. Let $\mathcal{V}=\mathcal{V}_{\mathbf{j}}\cup \mathcal{V}_{\mathbf{k}}$, Similarly, the edge sets describing the equality constraints between unanchored points and between anchor and unanchored points are $\mathcal{E}_\mathbf{j}\subseteq \mathcal{V}_{\mathbf{j}}\times \mathcal{V}_{\mathbf{j}}$ and $\mathcal{E}_\mathbf{k} \subseteq \mathcal{V}_{\mathbf{j}}\times \mathcal{V}_{\mathbf{k}}$ respectively. We represent the equality constraints with a weighted directed acyclic graph $\mathcal{G} = (\mathcal{V}, \mathcal{E}_{eq}, \ell)$, where $\ell\,:\, \mathcal{E}_{eq} \rightarrow {\rm I\!R}_{+}$ encodes the distances:
	\begin{align}
		\ell \,:\, (i,j) \mapsto || \mathbf{x}_i - \mathbf{x}_j ||
	\end{align}
	where $ \mathbf{x}_i $ and $\mathbf{x}_j$ can refer to either unanchored or anchored points, and $\mathcal{E}_{eq}=\mathcal{E}_{\mathbf{j}}\cup\mathcal{E}_{\mathbf{k}}$. The incidence matrix $\mathbf{B}(\mathcal{E}_{eq}) \in {\rm I\!R}^{|\mathcal{V}|\times|\mathcal{E}_{eq}|}$ can be used to compactly summarize the constraints (detailed in Section \ref{subsec:constr}), as
	\begin{align}
		\boldsymbol{\ell} &=\text{diag} \left( \mathbf{B}(\mathcal{E}_{eq})^T\mathbf{P}^T\mathbf{P}\mathbf{B}(\mathcal{E}_{eq})\right)  \label{eq:all_constr}\\
		\text{with }\mathbf{P} &= \left[ \begin{matrix} \mathbf{X} & \mathbf{W} \end{matrix}\right] \in  {\rm I\!R}^{d\times (j+k)}\\
		\mathbf{X} &= \left[ \begin{matrix} \mathbf{q}_1 &  \mathbf{p}_1 & \cdots & \mathbf{q}_n \end{matrix}\right] \in {\rm I\!R}^{d\times j}\\
		{\ell}_e &= \ell(e)^2  \enspace \forall e \in \mathcal{E}
	\end{align}
	This formulation is a modified form of \cite{giamou2022cidgik} and is equivalent to the one used in \cite{so2007theory} for SNL (sensor network localization). 
	
	\subsection{Constraints}
	\label{subsec:constr}
	
	The following constraints are used to form a QCQP that is equivalent to the IK problem for CRs.
	Each CC segment is indexed by $t$ and is associated with a nonnegative scalar variable $\omega_t$.
	
	\subsubsection{End-Effector and Base Specification} \label{sec:ee_specification}
\begin{figure}[]
		\centering
		\includegraphics[width=0.22\textwidth]{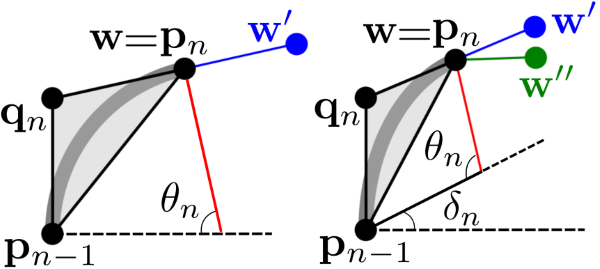}
		\caption{Anchors used to specify end-effector position/pose of an n-segment CR for planar (left) and spatial (right) IK.}
		\label{fig:singseg}
	\end{figure}

	End-effector position, orientation, and full pose targets are enforced via distance relationships between unanchored points and neighbouring anchors. 
	All anchor points $\mathbf{a}^\prime$ are set a unit distance away from their corresponding point $\mathbf{a}$, allowing us to specify a particular arrangement's orientation. 
	The robot base and end-effector orientation can be fully specified by:
	\begin{align}
		\mathbf{q}_{0} &= \mathbf{p}_0 + \omega_0(\mathbf{b}-\mathbf{b}^\prime) \label{eq:base}\\
		\mathbf{q}_n &= \mathbf{p}_n + \omega_n(\mathbf{p}_n-\mathbf{w}^\prime) \label{eq:ee_2}\\
		\| \mathbf{w}^{\prime\prime} - \mathbf{q}_n\|^2 &= \| \mathbf{w} - \mathbf{q}_n\|^2 +  \|\mathbf{w} - \mathbf{w}^{\prime\prime}\|^2 \label{eq:ee_3}\\
		\| \mathbf{w}^{\prime\prime} - \mathbf{p}_{n-1}\|^2 &= \| \mathbf{w} - \mathbf{p}_{n-1}\|^2 +  \|\mathbf{w} - \mathbf{w}^{\prime\prime}\|^2. \label{eq:ee_4}
	\end{align}
	With this formulation, it is possible to solve IK with a variety of EE specifications. 
	For the spatial case ($d=3$) these are: i) position only (inherent), ii) position, pitch, and yaw (Eq. \ref{eq:ee_2}), and iii) position, pitch, yaw, and roll (Eq. \ref{eq:ee_2}, \ref{eq:ee_3}, \ref{eq:ee_4}). 
	For the simpler planar case ($d=2$), i) and ii) are sufficient to constrain position and orientation, respectively (see Fig. \ref{fig:singseg}).
	Note that for $d=3$, the constraints in Eqs. \ref{eq:ee_2}, \ref{eq:ee_3}, and \ref{eq:ee_4} permit a \textit{reflection} about the $\mathbf{w}-\mathbf{w}^{\prime}-\mathbf{w}^{\prime\prime}$ plane, allowing us to achieve an EE specification just short of full six DoF.
	This enables us to almost fully specify the roll of the EE, whereas FABRIKc and its successors cannot specify the roll at all. Furthermore, many manipulators have a final joint with unlimited rotation along the roll axis, in which case this specification is sufficient. 

	\subsubsection{Body Constraints}
	\label{subsubsec:body_constr}
\begin{figure}[]
		\centering
		\includegraphics[width=0.35\textwidth]{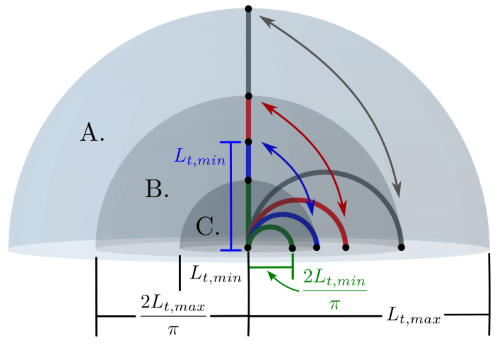}
		\caption{Shows workspace of an extensible single segment continuum robot and effect of the segment length constraint Eq. \ref{eq:length} limiting the reachable workspace. Shows four different segments in their fully straightened and $\theta_t = \pi/2$ bending configuration. The full reachable workspace of an extensible segment is encapsulated by the innermost (green) and outermost (grey) configurations. To guarantee our method returns physically realizable solutions, we limit the reachable workspace to one bounded by the two inner configurations, i.e. the blue and red segment configurations.}
		\label{fig:workspacelim}
	\end{figure}

	To ensure our method returns valid robot configurations (i.e., isosceles triangles encompassing each segment), the following constraints are used.\\
	\noindent\textbf{Symmetry}. This constraint enforces that linkages $l_t$ connecting segment endpoints to their respective virtual joints are equal-length:
	\begin{align}
		l_t = \| \mathbf{p}_{t-1} - \mathbf{q}_t\|^2 = \| \mathbf{p}_{t} - \mathbf{q}_t\|^2.
	\end{align}
	\noindent\textbf{Segment Continuity}. A smooth shape is enforced by requiring that a segment's virtual joint, end-point, and its succeeding virtual joint are collinear:
	\begin{align}
		\mathbf{q}_{t+1} = \mathbf{p}_t + \omega_t(\mathbf{p}_t-\mathbf{q}_t).
	\end{align}
	\textbf{Length}: Our model ensures that the scale of each segment triangle corresponds to a feasible solution for \textit{extensible} segment CR only. 
	The chord of each segment's CC arc is denoted $d_t \stackrel{\rm def}{=} \| \mathbf{p}_t - \mathbf{p}_{t-1} \|$ as shown Fig. \ref{fig:2d_model}c). 
	CR configurations with extreme bending are often unfavourable, therefore we restrict $\theta_t \in[0, \pi)$, Figs. \ref{fig:2d_model}a)-e), to omit such cases.
	For illustrative purposes, consider a \textit{fixed-length} segment CR for which $d_t$ falls in the range $d_t\in({2L_t}/{\pi}, L_t]$.
	A naive constraint proposal would be $\left(2L_t / \pi\right)^2 < d_t^2 \leq  \left(L_t\right)^2$. 
	The issue lies is that each bending angle $\theta_t$ has a \textit{unique} $l_t$ and $d_t$. This one-to-one mapping must be respected by the solver to ensure that solutions are physically realizable.
	Alone, this naive approach is not sufficient to enforce this mapping: a bent segment can be scaled to unrealizable sizes (i.e., segment lengths) within the range for $d_t$ while respecting all proposed constraints. For full-range \textit{extensible} segments, using an analogous length constraint will similarly result in physically unrealizable solutions. 
	\begin{align}
		\left(L_{t,min}\right)^2 \leq d_t^2 \leq \left({2L_{t,max}}/{\pi}\right)^2 \label{eq:length}
	\end{align}
	The constraint in Eq. \ref{eq:length} limits the full range of an extensible segment $L_t\in[L_{t, min}, L_{t,max}]$ in order to ensure that physically realizable solutions are returned despite this scaling limitation.
	Constraining a given segment's $d_t$ using Eq. \ref{eq:length} constrains the segment tip to remain in the shaded region B in Fig. \ref{fig:workspacelim}, effectively reducing the CR's workspace. 
	In many applications, this limitation is a useful feature, as it keeps the robot away from extremal configurations which correspond to kinematic singularities. 
	
	\subsubsection{Obstacle Avoidance}
	Consider a workspace $\mathcal{W}$ with obstacles or regions that the robot is forbidden from occupying. 
	We model these constraints with a finite set of spheres $\mathcal{O}$ whose union covers all restricted regions:
	\begin{align}
		||\mathbf{x}_i - \mathbf{c}_g||^2 \geq l_g^2 \,\,\,\forall i \in \mathcal{V}_{\mathbf{u}}, \forall g \in \mathcal{O} \label{eq:obs_constr},
	\end{align}
	where $\mathbf{c}_g\in{\rm I\!R}^d$ is the center, $\mathbf{x}_i$ are the unanchored segment base and tip points $\mathbf{p}_t$ and $\mathbf{p}_{t-1}$, and $l_g > 0$ is the radius of sphere $g \in \mathcal{O}$. 
	This union of spheres environment representation can represent complex obstacles up to arbitrary precision when a large number of spheres is used. 
	To specify a spherical region of free space \textit{within} which some subset of unanchored points must lie, the inequality in Eq. \ref{eq:obs_constr} is reversed. 
	Additionally, it is possible to constrain a point $\mathbf{x}\in{\rm I\!R}^d$ on the robot to lie on one side of a plane using a linear inequality constraint with the plane's normal $\mathbf{n}$ and its minimum Euclidean distance $c$ from the origin: 
	\begin{align}
		\mathbf{x}^T\mathbf{n} \leq c \label{eq:half_plane}
	\end{align}
	
	\subsection{QCQP Formulation}
	\label{subsec:qcqp_form}
	We use the quadratic and linear expressions of Eq. \ref{eq:all_constr}, \ref{eq:obs_constr}, and \ref{eq:half_plane} to describe solutions to IK as the following quadratic feasibility program:
	
	\textit{Problem 2} (Feasibility QCQP)
	\begin{align*}
		\text{find} \, &\mathbf{X} \in {\rm I\!R}^{d\times j}\\
		\text{s.t.} \quad &\mathbf{P} = \left[ \begin{matrix} \mathbf{X} & \mathbf{B} & \mathbf{W} \end{matrix}\right], \\
		&\text{diag} \left( \mathbf{B}(\mathcal{E}_{eq})^T\mathbf{P}^T\mathbf{P}\mathbf{B}(\mathcal{E}_{eq})\right) = \mathbf{d},\\
		&||\mathbf{x}_i - \mathbf{c}_g||\geq l_g^2 \,\; \forall i \in \mathcal{V}_{\mathbf{u}}, \; \forall g \in \mathcal{O},\\
		&\mathbf{x}_i^T\mathbf{n}_i \leq c_i \; \forall i \in \mathcal{V}_{\mathbf{p}} \subset \mathcal{V}_{\mathbf{u}},
	\end{align*}
	\textit{where $\mathcal{E}_{eq}$, $\mathbf{d}$, $\mathbf{B}$, $\mathbf{W}$ are problem parameters, $\mathcal{O}$ is the set of spherical obstacle constraints of the form in Eq. \ref{eq:obs_constr}, and $\mathcal{V}_{\mathbf{p}}$ indexes the nodes confined to lie in some plane described by $\mathbf{n}_i$ and $\mathbf{n}_i$ in Eq. \ref{eq:half_plane}.}
	
	\section{Algorithm}
	\label{sec:algo}
	When solutions exist, the feasible set of Problem 2 is nonconvex and therefore challenging to characterize. This section outlines our IK solver which leverages semidefinite programming (SDP) relaxations as a means to efficiently compute solutions to Problem 2. Similar to \cite{so2007theory} and \text{CIDGIK} \cite{giamou2022cidgik}, at the core of \texttt{CIDGIKc} is our modified form of the convex iteration algorithm from Dattoro \cite{dattorro2010convex} summarized below.
	\subsection{Semidefinite Relaxations}
	\label{subsec:sdp_relax}
	Semidefinite programming (SDP) relaxations are used as a means of efficiently computing solutions to Problem 2. The lifted matrix variable is defined as
	\begin{align}
		\mathbf{Z}(\mathbf{X}) &\stackrel{\rm def}{=} \left[\begin{matrix}\mathbf{X}&\mathbf{\Omega}&\mathbf{I}_d\end{matrix}\right]^T\left[\begin{matrix}\mathbf{X}&\mathbf{\Omega}&\mathbf{I}_d\end{matrix}\right] 
		\\&= \left[\begin{matrix}\mathbf{X}^T\mathbf{X} & \mathbf{X}^T\mathbf{\Omega} & \mathbf{X}^T \\ \mathbf{\Omega}^T\mathbf{X} & \mathbf{\Omega}^T\mathbf{\Omega}  & \mathbf{\Omega}^T\\ \mathbf{X} & \mathbf{\Omega} & \mathbf{I}_d\end{matrix}\right] \in \mathbf{S}_+^{m}\label{eq:lifting}
	\end{align}
	where $\mathbf{\Omega} = \left[\begin{matrix}\omega_0\mathbf{I}_d & \dots & \omega_t\mathbf{I}_d & \omega_{\mathbf{w}^\prime}\mathbf{I}_d \end{matrix}\right]$. 
	For the cases in  Fig. \ref{fig:2d_model}b) and \ref{fig:2d_model}c), $\omega_{\mathbf{w}^\prime}\mathbf{I}_d$ is included such that $\mathbf{\Omega} \in {\rm I\!R}^{(n+1)d}$, and ${\rm I\!R}^{nd}$ otherwise. 
	Similarly, for EE specifications ii) and iii) from Sec. \ref{sec:ee_specification}, $m =	 j + (n+2)d$, and $m = j + (n+1)d$ otherwise. 
	This allows us to rewrite the quadratic constraints of \textit{Problem 2} as \textit{linear} functions of $\mathbf{Z}(\mathbf{X})$. Since $\mathbf{Z}(\mathbf{X})$ is an outer product of matrices with rank of at most $d$ (the dimension of the space in which the robot operates), we know that $rank(\mathbf{Z})\leq d$. Replacing $\mathbf{Z}(\mathbf{X})$ with a \textit{general} PSD matrix $\mathbf{Z} \succeq \mathbf{0}$ produces the following semidefinite relaxation:
	\textit{Problem 3} (SDP Relaxation of \textit{Problem 2})
	\begin{align*}
		\text{find} \, &\mathbf{Z} \in  \mathbb{S}^m_+ \\
		\text{s.t.} \quad &\mathcal{A}(\mathbf{Z}) = \mathbf{a},\\
		& \mathcal{B}(\mathbf{Z}) \leq \mathbf{b},
	\end{align*}
	\textit{where $\mathcal{A}: \mathbb{S}^{m}_+ \rightarrow {\rm I\!R}^{k+d^{2}+|\mathcal{V}|}$ and $\mathbf{a} \in {\rm I\!R}^{k+d^{2}+|\mathcal{V}|}$ encode the linear equations that enforce the constraints in Eq. \ref{eq:all_constr} and Eq. \ref{eq:half_plane} after applying the substitution in Eq. \ref{eq:lifting}, and the linear map $\mathcal{B}:\mathbb{S}^{m}_+ \rightarrow {\rm I\!R}^{|\mathcal{O}|}$ with vector $\mathbf{b} \in {\rm I\!R}^{|\mathcal{O}|}$ enforces the inequalities in Eq. \ref{eq:obs_constr}.}
	
	Problem 3 is a convex feasibility problem, which can be solved by interior-point methods. 
	However due to the nature of the relaxation, solutions to Problem 3 are not limited to proper rank-$d$ solutions originally sought after in \textit{Problem 2} (see \cite{giamou2022cidgik} for further explanation). 
	
	\subsection{Rank Minimization} 
	As in \cite{giamou2022cidgik}, we choose to minimize convex (linear) heuristic cost functions that encourage low rank solutions:
	
	\textit{Problem 4} (Problem 3 with a Linear Cost $\mathbf{C}\in \mathbb{S}^m_+$) \textit{Find the symmetric PSD matrix $\mathbf{Z}$ that solves}
	\begin{align}
		\min_{\mathbf{Z} \in \mathbb{S}^m_+} \quad &\text{tr}(\mathbf{C}\mathbf{Z}) \\
		\text{s.t.} \quad &\mathcal{A}(\mathbf{Z}) = \mathbf{a}, \\
		& \mathcal{B}(\mathbf{Z}) \leq \mathbf{b}. \label{eqn:p4}
	\end{align}
	\textit{Problem 5} (Sum of $2n$ Smallest Eigenvalues as SDP) \textit{Find the symmetric PSD matrix $\mathbf{C}$ that solves}
	\begin{align}
		\sum\limits_{i=d+1}^{m} \lambda_i(\mathbf{Z}) = \min_{\mathbf{C} \in \mathbb{S}^m_+}  \quad & \text{tr}(\mathbf{C}\mathbf{Z}), \\
		\text{s.t.} \quad &\text{tr}(\mathbf{C}) = 2n,\\
		& \mathbf{0} \preccurlyeq \mathbf{C} \preccurlyeq \mathbf{I}. \label{eqn:p5}
	\end{align}
	
	\subsection{Convex Iteration} \label{sec:convex_iteration}
	
	The convex iteration algorithm \cite{dattorro2010convex} alternates between solving Problem 4 and Problem 5 to find feasible low-rank solutions to Problem 2. 
	We fond initializing \textit{Problem 4} with a $\mathbf{Z}$ reconstructed from a configuration with all segments straight and at mid-extension greatly reduced the iterations required to converge. We opted to incorporate this modification into \texttt{CIDGIKc}, which is summarized in Algorithm \ref{alg:convex_iteration}. 
	Since \textit{Problem 5} has a fast closed-form solution \cite{dattorro2010convex}, most of \texttt{CIDGIKc}'s computational cost comes from solving \textit{Problem 4}.
	
	\begin{algorithm}
		\caption{Convex Iteration for Distance Geometric IK for Continuum Robots \texttt{CIDGIKc}}\label{alg:convex_iteration}
		\begin{algorithmic}
			\State \textbf{Input:} \textit{Problem 4} specification (i.e. $\mathcal{A}, \mathcal{B}, \mathbf{a}, \mathbf{b}$)
			\State \textbf{Result:} PSD matrix $\mathbf{Z}^\star$ that solves \textit{Problem 4}
			\State Warm start with $\mathbf{Z}^{\{i\}}$ as a rank-$d$ solution corresponding to all segments straight and at mid-extension
			\State Get the initialization for $\mathbf{C}^{\{i\}}$ using \textit{Problem 5} and warm start $\mathbf{Z}^{\{i\}}$
			\While{\textbf{\textit{not converged} do}} 
			\State Solve $\mathbf{Z}^{\{i\}}=\text{argmin}_{\mathbf{Z}}$ \textit{Problem 4} with $\mathbf{C} = \mathbf{C}^{\{i\}}$
			\State Solve $\mathbf{C}^{\{i\}}=\text{argmin}_{\mathbf{C}}$ \textit{Problem 5} with $\mathbf{Z} = \mathbf{Z}^{\{i\}}$
			\EndWhile
			\State Return $\mathbf{Z}^\star = \mathbf{Z}^{\{i\}}$
		\end{algorithmic}
		\label{algo:cidgk}
	\end{algorithm}
	
	
	\section{Experiments}
	\label{sec:exp}
\begin{figure*}[t!]
		\centering
		\begin{subfigure}{0.18\textwidth}
			\centering
			\includegraphics[width=\textwidth]{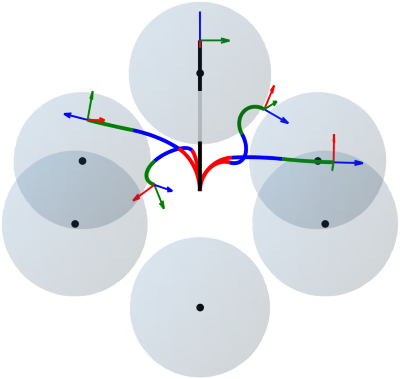}
			\caption{\textit{Octahedron}(6).}
			\label{fig:obs_env_0}
		\end{subfigure}
		\begin{subfigure}{0.17\textwidth}
			\centering
			\includegraphics[width=\textwidth]{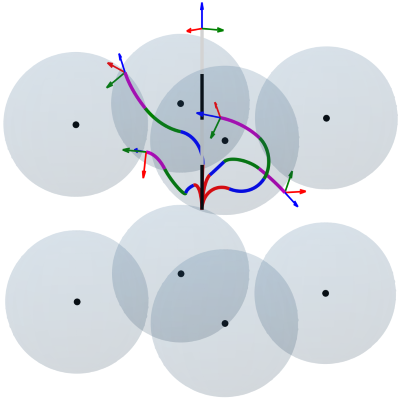}
			\caption{\textit{Cube} (8).}
			\label{fig:obs_env_1}
		\end{subfigure}
		\begin{subfigure}{0.16\textwidth}
			\centering
			\includegraphics[width=\textwidth]{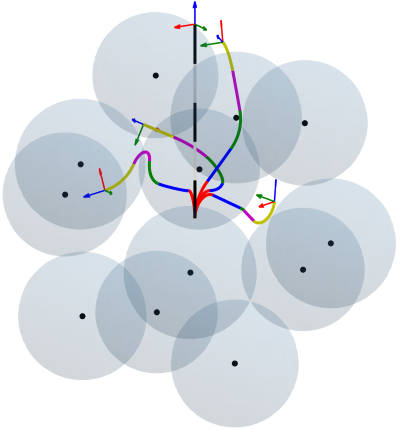}
			\caption{\textit{Icosahedron} (12).}
			\label{fig:obs_env_2}
		\end{subfigure}
		\begin{subfigure}{0.24\textwidth}
			\centering
			\includegraphics[width=\textwidth]{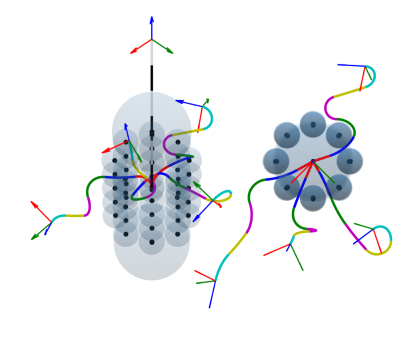}
			\caption{\textit{Columns} (42).}
			\label{fig:obs_env_3}
		\end{subfigure}
		\begin{subfigure}{0.22\textwidth}
			\centering
			\includegraphics[width=\textwidth]{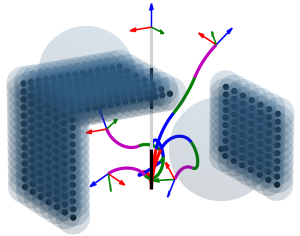}
			\caption{\textit{Corridor} (261).}
			\label{fig:obs_env_4}
		\end{subfigure}
		\caption{Five spatial obstacle configurations, environments, (number of spherical obstacles in `()') used in to test \texttt{CIDGIKc}. The environments (a) to (c) correspond to obstacle configurations defined by the vertices of the respective platonic solid centered at the base of the robot. The straight initial configuration and four example robot configurations are shown in each scenario for illustrative purposes.}
		\label{fig:obs_env_all}
	\end{figure*}

	We evaluate our \texttt{CIDGIKc} for realistic planar and spatial continuum robots, $n=3,4,5,6$, operating within the cluttered environments shown in Fig. \ref{fig:obs_env_all}. 
	Environment geometries (obstacle sizes and distances from one another) are scaled proportionately with respect to $n$. 
	All segments have $L_{t,min} = 0.15\text{ m}$, $L_{t,max} = 0.55\text{ m}$, the same extensible range as \cite{burgnerkhars2021tendon}. We generate \textit{feasible} IK problem instances by sampling the forward kinematics $\mathbf{\Theta} \in \mathbf{\mathcal{C}}$, and use the resulting EE pose $F(\Theta)$ as a query position or pose goal. 
	Configurations with self-collisions, collision with environmental obstacles (if present), and any part of the robot body below the base are eliminated. 
	IK queries are generated with arc parameters $\theta_t \in [0^\circ, 179.5^\circ]$ and $\delta_t \in [0^\circ, 360^\circ]$ drawn from a random uniform distribution and $L_t$ drawn from a normal distribution with $\mu = 0.35\text{ m}$ and $\sigma = 0.075\text{ m}$ such that $L_t \in [L_{t,min}, L_{t,max}]$. 
	Each problem instance is solved with Algorithm \ref{alg:convex_iteration} (\texttt{CIDGIKc}) using the MOSEK interior point solver \cite{andersen2000mosek} for each iteration of \textit{Problem 4}. 
	Solutions are recovered by reconstructing the configuration $\Theta$ from points $\mathbf{X}$ extracted from $\mathbf{Z}^\star$ returned after the $(d+1)^{\text{th}}$ eigenvalue of $\mathbf{Z}$ drops below $10^{-7}$, indicating convergence to a rank-$d$ solution, or after 200 iterations have elapsed. 
	Our solution recovery strategy is identical to other point-based IK methods leveraging the CC assumption (see \cite{zhang2018fabrikc}). 
	The EE pose and workspace constraint violations are computed by running the forward kinematics on the recovered $\Theta$. 
	A solution is considered \textit{valid} when obstacle avoidance with segment endpoints is achieved to within a 0.01 m tolerance, there are no self-collisions, segment lengths are within the permissible range, EE position error is below 1\% of the full robot length with each segment at mid-extension, and EE rotation error is below 2$^\circ$. 
	Rotation errors are computed as the angle between the desired and actual $z, y$ pose vectors, accounting for reflections (see \ref{subsec:constr}). 
	
	We saw improved solver performance (convergence time and success) by excluding the upper bound on $d_t$ for Eq. \ref{eq:length} in all cases except for specification i). 
	Without explicitly bounding $d_t$ from above, starting from a valid robot initialization (see Section \ref{sec:algo}) still leads to $>$95\% valid segment lengths across all solves.
	Intuitively, with EE specifications ii) and iii), additional orientation constraints offer more information as to the underlying structure and thus require fewer constraints. 
	\texttt{CIDGIKc} was tested on a variety of test cases considering different $n$, EE specifications, and environments. 
	The initialization described in Sec. \ref{sec:convex_iteration} is used unless otherwise stated. 
	All experiments were implemented in Python and run on a PC with an Intel i7-9800X CPU running at 3.80 GHz and with 16 GB of RAM. 
	We define solution time as the sum of solver times over all iterations (performing \textit{Problem 4} and \textit{5}), without measuring the setup time of each algorithm for \texttt{CIDGIKc}.
	
\begin{figure}[]
		\centering
		\includegraphics[width=0.45\textwidth]{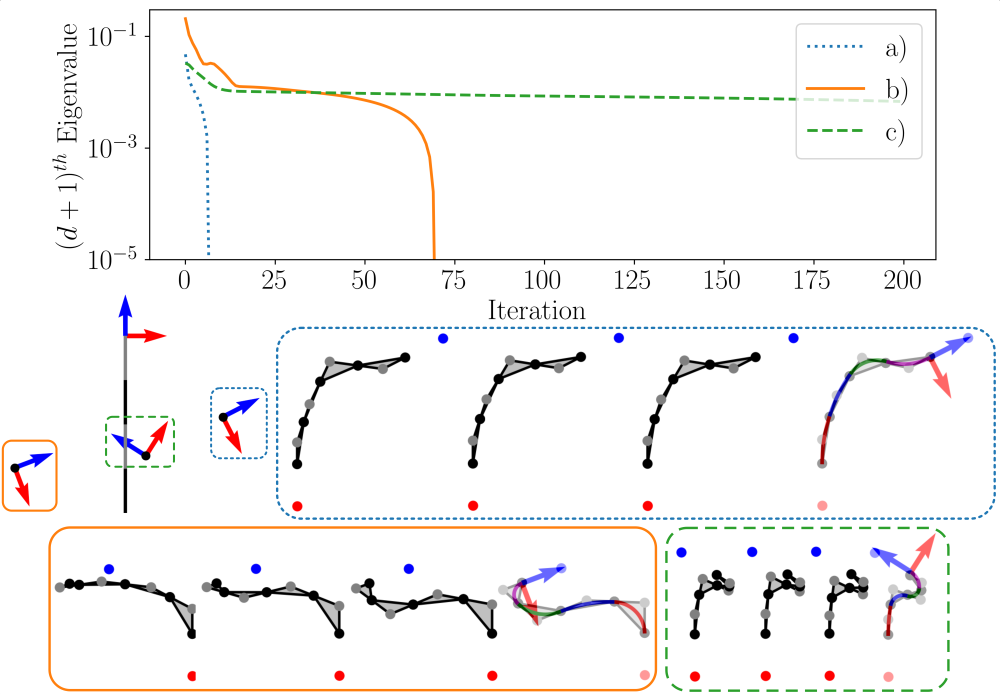}
		\caption{Three \texttt{CIDGIKc} solves of a planar four segment extensible robot. Top plot shows the $(d+1)^{\text{th}}$ eigenvalue of the $\mathbf{Z}$ matrix until convergence of IK solves given EE position and orientation queries a)-c) shown in the middle-left plot with the straight initialization. Each query took 8, 65, and 200 iterations to converge respectively. Below are snapshots of the solving process for each query shown as 3 equally-spaced intermediate recovered solutions and the final recovered solution.}
		\label{fig:representative_cases}
	\end{figure}

	We generated 400 full pose IK queries in free space consisting of 100 queries for $n=3,4,5,6$. For each query, the full range of EE specifications were obtained from the full pose.
	As per Table \ref{tab:eespec_results}, $>$98\% of solves converged and $>$95\% of them were considered valid, demonstrating the expected functionality of the different EE specifications.
	Representative $d=2$ solves are shown in Fig. \ref{fig:representative_cases}. 
	Note the behaviour for problem c), where despite not converging after $k_{max}$ iterations, a reasonable approximate solution is recovered. 
	
	\begin{table}[]
		\centering
		\footnotesize
		\setlength{\tabcolsep}{1pt}
		\caption{Average end-effector specification errors.}\label{tab:eespec_results}
		\resizebox{0.485\textwidth}{!}{
			\begin{tabular}{llllll}
				& pos. err. [m] & rot. $z$ err. [\textdegree] & rot. $y$ err. [\textdegree] & $\%$ con. & $\%$ val.\\ \midrule\midrule
				\multirow{2}{*}{$d=2$ i) pos.}  & 3.6$\times10^{-8}$ & 4.2$\times10^{1}$ & & \multirow{2}{*}{100 $\pm$ 0.0} & \multirow{2}{*}{94.5 $\pm$ 2.2} \\
				& $\pm$ 5.1$\times10^{-8}$  & $\pm$ 2.6$\times10^{1}$ & & & \\ 
				\multirow{2}{*}{$d=2$ ii) pos. + y,p}  & 5.0$\times10^{-8}$ & 3.5$\times10^{-7}$ & & \multirow{2}{*}{98 $\pm$ 1.4} & \multirow{2}{*}{95 $\pm$ 2.1} \\
				& $\pm$ 8.7$\times10^{-8}$  & $\pm$ 7.8$\times10^{-7}$ & & & \\
				\midrule
				\multirow{2}{*}{$d=3$ i) pos.}  & 4.6$\times10^{-8}$ & 5.7$\times10^{1}$ & 5.3$\times10^{1}$ & \multirow{2}{*}{100 $\pm$ 0.0} & \multirow{2}{*}{96.5 $\pm$ 1.8} \\
				& $\pm$ 6.5$\times10^{-8}$  & $\pm$ 2.2$\times10^{1}$ & $\pm$ 2.3$\times10^{1}$ & & \\ 
				\multirow{2}{*}{$d=3$ ii) pos. + y,p}  & 5.4$\times10^{-8}$ & 3.9$\times10^{-7}$ & 3.9$\times10^{1}$ & \multirow{2}{*}{98.8 $\pm$ 1.1} & \multirow{2}{*}{96.0 $\pm$ 1.9} \\
				& $\pm$ 7.6$\times10^{-8}$  & $\pm$ 9.4 $\times10^{-7}$ & 2.7 $\times10^{1}$ & & \\
				\multirow{2}{*}{$d=3$ iii) pos. + y,p,r}  & 4.9$\times10^{-8}$ & 4.1$\times10^{-7}$ & 3.4$\times10^{-7}$ & \multirow{2}{*}{98.3 $\pm$ 1.3} & \multirow{2}{*}{95.0 $\pm$ 2.1} \\
				& $\pm$ 7.2$\times10^{-8}$  & $\pm$ 8.8$\times10^{-7}$ & $\pm$ 7.5$\times10^{-7}$ & & \\ \hline
			\end{tabular}
		}
		\vspace{0.1cm}
		
		Abbreviations: y=yaw, p=pitch, r=roll, con.=converged, val=valid.
	\end{table}
	
	\begin{figure}[]
		\centering
		\includegraphics[width=\columnwidth]{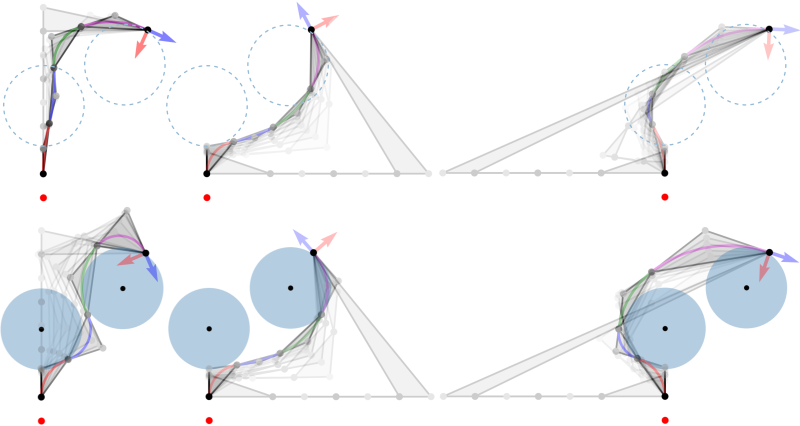}
		\caption{Position only IK solves using \texttt{CIDGIKc} for a planar, four segment, continuum robot without (top) and with (bottom) obstacles considering a straight (right), right sideways (middle), and left sideways (left) initialization. Initializations and equally-spaced recovered solutions throughout the solves are shown in light grey with the final recovered solution body shown in color.}
		\label{fig:obstacle_case}
	\end{figure}
	\begin{table*}
		\footnotesize
		\centering
		\caption{Percentages of successfully solved problems as 95\% Jeffreys confidence intervals.}\label{tab:success_nums}
		\resizebox{0.98\textwidth}{!}{
			\begin{tabular}{lcccccccccc}
				Env. & \multicolumn{2}{c}{Octahedron} & \multicolumn{2}{c}{Cube} & \multicolumn{2}{c}{Icosahedron}& \multicolumn{2}{c}{Columns}& \multicolumn{2}{c}{Corridor}\\
				\cmidrule(r){2-3} \cmidrule(r){4-5}  \cmidrule(r){6-7} \cmidrule(r){8-9}  \cmidrule(r){10-11}
				&\texttt{free}&\texttt{with obs} &\texttt{free}&\texttt{with obs}&\texttt{free} &\texttt{with obs}&\texttt{free}&\texttt{with obs} &\texttt{free}&\texttt{with obs}\\
				\midrule
				\midrule
				$n=3$ &	97.6	 $\pm$	1.9	&	96.8	 $\pm$	2.2	&	99.2	 $\pm$	1.1	&	99.2	 $\pm$	1.1	&	99.2	 $\pm$	1.1	&	99.2	 $\pm$	1.1	&	99.2	 $\pm$	1.1	&	98.8	 $\pm$	1.3	&	100.0	 $\pm$	0.0	&	100.0	 $\pm$	0.0	\\
				$n=4$ 	&	99.6	 $\pm$	0.8	&	98.8	 $\pm$	1.3	&	99.2	 $\pm$	1.1	&	98.8	 $\pm$	1.3	&	100.0	 $\pm$	0.0	&	99.6	 $\pm$	0.8	&	100.0	 $\pm$	0.0	&	99.6	 $\pm$	0.8	&	100.0	 $\pm$	0.0	&	100.0	 $\pm$	0.0\\
				$n=5$ &	99.6	 $\pm$	0.8	&	99.6	 $\pm$	0.8	&	100.0	 $\pm$	0.0	&	100.0	 $\pm$	0.0	&	100.0	 $\pm$	0.0	&	100.0	 $\pm$	0.0	&	100.0	 $\pm$	0.0	&	99.6	 $\pm$	0.8	&	100.0	 $\pm$	0.0	&	99.6	 $\pm$	0.8\\
				$n=6$ &	100.0	 $\pm$	0.0	&	99.2	 $\pm$	1.1	&	99.2	 $\pm$	1.1	&	99.2	 $\pm$	1.1	&	98.8	 $\pm$	1.3	&	98.4	 $\pm$	1.6	&	99.6	 $\pm$	0.8	&	98.8	 $\pm$	1.3	&	99.6	 $\pm$	0.8	&	99.6	 $\pm$	0.8\\
			\end{tabular}
		}
		\vspace{0.25cm}
		\footnotesize
		\centering
		\caption{Mean, with standard deviation, of iterations used.}\label{tab:its_nums}
		\resizebox{0.98\textwidth}{!}{
			\begin{tabular}{lcccccccccc}
				Env. & \multicolumn{2}{c}{Octahedron} & \multicolumn{2}{c}{Cube} & \multicolumn{2}{c}{Icosahedron}& \multicolumn{2}{c}{Columns}& \multicolumn{2}{c}{Corridor}\\
				\cmidrule(r){2-3} \cmidrule(r){4-5}  \cmidrule(r){6-7} \cmidrule(r){8-9}  \cmidrule(r){10-11}
				&\texttt{free}&\texttt{with obs} &\texttt{free}&\texttt{with obs}&\texttt{free} &\texttt{with obs}&\texttt{free}&\texttt{with obs} &\texttt{free}&\texttt{with obs}\\
				\midrule
				\midrule
				$n=3$ &	27.7	 $\pm$	38.0	&	27.5	 $\pm$	37.9	&	29.8	 $\pm$	46.4	&	29.8	 $\pm$	46.4	&	28.8	 $\pm$	41.2	&	28.8	 $\pm$	41.2	&	20.8	 $\pm$	36.8	&	21.9	 $\pm$	36.8	&	29.0	 $\pm$	43.2	&	29.4	 $\pm$	43.2\\
				$n=4$ &	35.9	 $\pm$	28.0	&	38.6	 $\pm$	29.5	&	35.1	 $\pm$	30.9	&	34.9	 $\pm$	31.0	&	32.1	 $\pm$	27.3	&	31.8	 $\pm$	26.9	&	22.6	 $\pm$	17.3	&	25.2	 $\pm$	23.9	&	29.0	 $\pm$	24.2	&	32.6	 $\pm$	29.2\\
				$n=5$ &	55.3	 $\pm$	40.0	&	59.2	 $\pm$	39.8	&	63.3	 $\pm$	40.8	&	63.3	 $\pm$	40.8	&	66.7	 $\pm$	39.4	&	68.4	 $\pm$	40.9	&	35.9	 $\pm$	25.9	&	38.3	 $\pm$	27.9	&	57.8	 $\pm$	39.5	&	64.2	 $\pm$	44.9\\
				$n=6$ &	71.8	 $\pm$	47.7	&	75.8	 $\pm$	49.2	&	80.6	 $\pm$	50.1	&	49.7	 $\pm$	50.2	&	94.0	 $\pm$	52.1	&	94.3	 $\pm$	52.1	&	56.9	 $\pm$	38.3	&	57.2	 $\pm$	37.1	&	86.3	 $\pm$	47.8	&	88.3	 $\pm$	48.7\\
			\end{tabular}
		}
		\vspace{0.25cm}
		\footnotesize
		\centering	
		\caption{Mean, with standard deviation, of solution times in seconds.}\label{tab:times_nums}
		\resizebox{0.98\textwidth}{!}{
			\begin{tabular}{lcccccccccc}
				Env. & \multicolumn{2}{c}{Octahedron} & \multicolumn{2}{c}{Cube} & \multicolumn{2}{c}{Icosahedron}& \multicolumn{2}{c}{Columns}& \multicolumn{2}{c}{Corridor}\\
				\cmidrule(r){2-3} \cmidrule(r){4-5}  \cmidrule(r){6-7} \cmidrule(r){8-9}  \cmidrule(r){10-11}
				&\texttt{free}&\texttt{with obs} &\texttt{free}&\texttt{with obs}&\texttt{free} &\texttt{with obs}&\texttt{free}&\texttt{with obs} &\texttt{free}&\texttt{with obs}\\
				\midrule
				\midrule
				$n=3$ &	3.1	 $\pm$	3.6	&	3.0	 $\pm$	3.5	&	3.2	 $\pm$	4.2	&	3.2	 $\pm$	4.2	&	3.2	 $\pm$	3.8	&	3.3	 $\pm$	3.9	&	2.5	 $\pm$	3.5	&	2.5	 $\pm$	3.2	&	3.0	 $\pm$	3.7	&	4.0	 $\pm$	4.4\\
				$n=4$ &	6.2	 $\pm$	4.1	&	6.6	 $\pm$	4.4	&	6.0	 $\pm$	4.5	&	5.8	 $\pm$	4.4	&	5.5	 $\pm$	3.9	&	5.6	 $\pm$	3.9	&	3.9	 $\pm$	2.3	&	4.6	 $\pm$	3.4	&	4.9	 $\pm$	3.4	&	7.3	 $\pm$	4.9\\
				$n=5$ &	14.2	 $\pm$	9.2	&	15.0	 $\pm$	9.1	&	16.4	 $\pm$	9.6	&	16.1	 $\pm$	9.4	&	17.1	 $\pm$	9.2	&	17.3	 $\pm$	9.5	&	9.3	 $\pm$	5.6	&	10.3	 $\pm$	6.2	&	14.1	 $\pm$	8.6	&	19.9	 $\pm$	11.9\\
				$n=6$ &	31.4	 $\pm$	19.1	&	20.1	 $\pm$	20.2	&	33.3	 $\pm$	19.1	&	34.8	 $\pm$	20.0	&	40.4	 $\pm$	21.1	&	40.7	 $\pm$	21.2	&	24.2	 $\pm$	14.5	&	26.2	 $\pm$	15.1	&	36.7	 $\pm$	18.9	&	46.6	 $\pm$	23.1\\
			\end{tabular}
		}
	\end{table*}
	250 full pose IK queries were generated for each $n=3,4,5,6$ within four distinct environments shown in Fig. \ref{fig:obs_env_all} with 6, 8, 12, 42, 261 spherical obstacles respectively. 
	The same queries were solved with (\texttt{with obs}) and without the obstacles (\texttt{free}) to provide direct performance comparisons shown in Tables \ref{tab:success_nums}-\ref{tab:times_nums}. 
	For environments with obstacles, \texttt{CIDGIKc} finds a \textit{valid} IK solution in 99.2\% of cases. 
	As expected, increasing the number of segments proportionately increases solver effort (i.e. iterations to convergence and solve times). 
	The \texttt{free} and \texttt{with obs} cases only differ slightly in terms of solver effort required. 
	In fact, contrary to intuition, there are cases where \texttt{with obs} requires less computation. There seems to be no relationship between solver effort and the number of obstacles considered. 
	It seems that solve time is not dictated by the number of obstacles but by the difficulty in solving the base \texttt{free} problem. 
	Planar solves in a contrived two-obstacle environment are shown in Fig. \ref{fig:obstacle_case} starting from straight, right, and left initializations at mid-extension, demonstrating that the robot can start at invalid colliding configurations and that the initialization can influence which of the redundant solutions is output. 
	\begin{figure}[]
		\centering
		\includegraphics[width=0.5\textwidth]{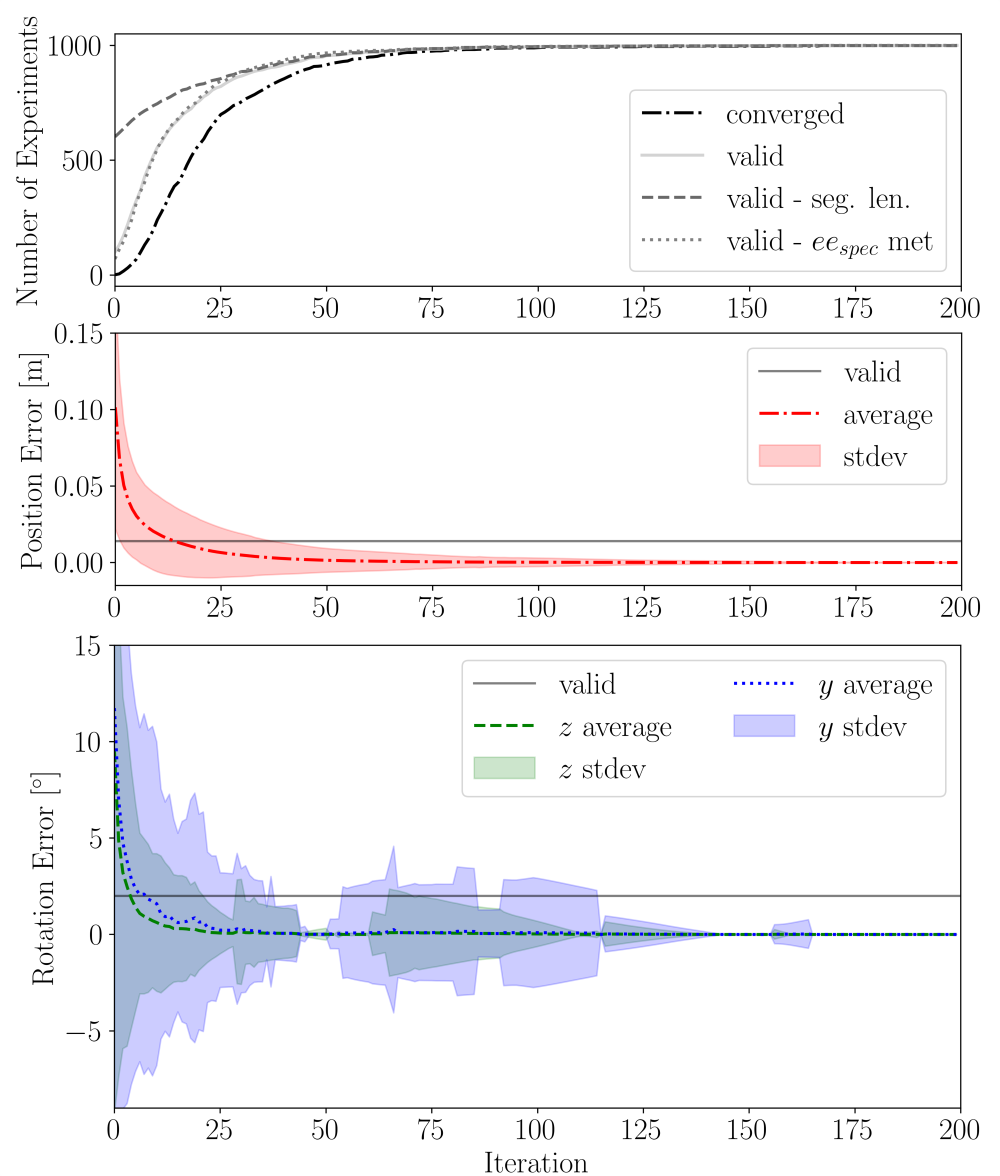}
		\caption{Averaged results from 1000 runs on full pose IK queries for a four segment spatial extensible continuum robot. Top plot shows in aggregate the number of experiments that converged or were valid at all iterations. Middle and bottom plots shows the average and standard deviation of end-effector position and orientation error performances respectively at all iterations.}
		\label{fig:indepth}
	\end{figure}
	\begin{figure}[]
		\centering
		\includegraphics[width=0.49\textwidth]{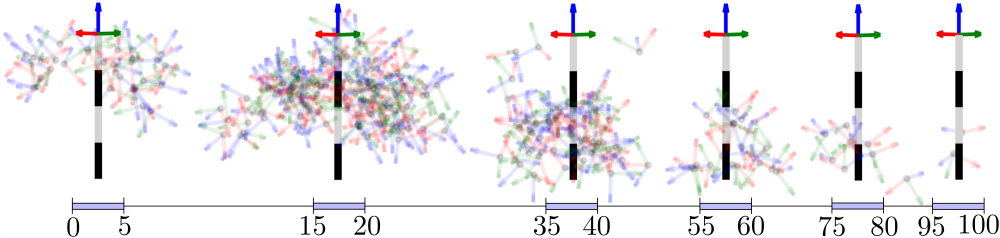}
		\caption{Full pose IK queries that converged within the range of iterations specified in the bottom scale bar. Straight mid-extension initialization shown for reference. Data from the experiment in Fig. \ref{fig:indepth}.}
		\label{fig:solve_difficulty}
	\end{figure}
	\begin{figure}[]
		\centering
		\includegraphics[width=0.49\textwidth]{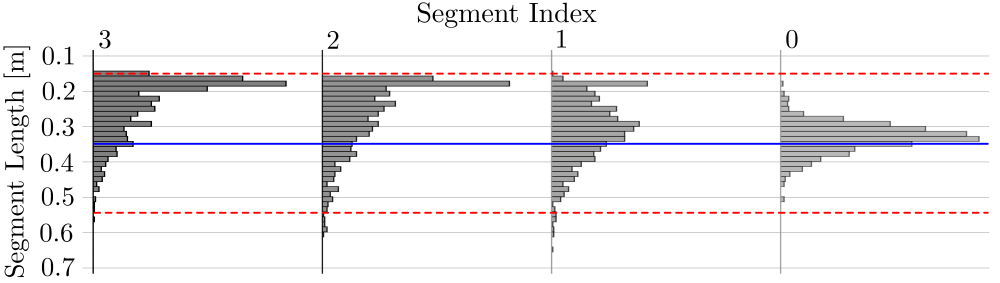}
		\caption{Histogram of \texttt{CIDGIKc} solution segment lengths, same data from Fig. \ref{fig:indepth}.}
		\label{fig:seg_length_distrib}
	\end{figure}
	
	Finally, 1000 full pose IK queries for a $n=4$, $d=3$ robot in free space were run through \texttt{CIDGIKc} with intermediate solutions recovered at each iteration to study how the solver approached solutions in terms of validity (valid segment lengths and EE specifications met) vs convergence shown in Fig. \ref{fig:indepth}. Within the first 30 iterations the majority of problems have rapidly approached or have met the EE specification and have valid segment lengths. The most challenging queries appear close to the segment base and far from the initial configuration as shown in Fig. \ref{fig:solve_difficulty}. 
	The segment lengths of output IK solutions are shown in Fig. \ref{fig:seg_length_distrib}. 
	\section{Conclusions and Future Work}
	\label{sec:conclusions}
	We have presented \texttt{CIDGIKc}, a novel distance-geometric approach to solving inverse kinematics problems involving extensible continuum robots that can handle position, partial, and full pose EE queries. 
	We demonstrated high efficacy ($>$99.2\%) in solving IK problems within a variety of cluttered environments and show that solve times don't increase with the number of obstacles considered. Crucially, our problem formulation connects IK to the rich literature on SDP relaxations for distance geometry problems, providing us with a novel and elegant geometric interpretation of IK.
	Our problem formulation connects IK for CRs to SDP relaxations for DGPs, enabling a variety of potential approaches. 
	Further work regarding the model will involve developing obstacle avoidance with the entire robot body (not just segment endpoints), and incorporating arbitrary joint angle limits. 
	Another area worth investigating is other methods to limit segment lengths to a permissible range. This may involve alternative length constraints or a post-processing step on the output of \texttt{CIDGIKc} that corrects near-valid segment length solutions produced by the solver. 
	There is also potential to speed up \texttt{CIDGIKc} by experimenting with initializations, implementing an SDP solver that exploits chordal sparsity, trying other DGP-compatible solving schemes (e.g., Riemannian optimization \cite{maric2021riemannian}), and producing an implementation in a compiled language. 
	Much like its serial arm counterpart \texttt{CIDGIK} \cite{giamou2022cidgik}, while our method uses global optimization to solve the subproblem in each iteration, there are no global convergence guarantees for the entire procedure. 
	These guarantees may depend on the precise problem formulation or hyperparameters which influence the particular feasible solutions \texttt{CIDGIKc} returns.
	
	\bibliographystyle{IEEEtran}
	\bibliography{IEEEabrv,references}
	
\end{document}